\documentclass[runningheads]{llncs}

\usepackage[letterpaper,top=2.25cm,bottom=2.25cm,left=2.4cm,right=2.4cm,marginparwidth=1.75cm]{geometry}
\usepackage[english]{babel}
\usepackage{graphicx}
\usepackage{csquotes}
\usepackage{dirtytalk}

\usepackage{titlesec}
\titlespacing*{\section}{0pt}{1.0\baselineskip}{0.8\baselineskip}
\titlespacing*{\subsection}{0pt}{0.8\baselineskip}{0.4\baselineskip}

\usepackage[
    style=apa,
    backend=biber,
    sortcites=true,
    sorting=nyt,
    hyperref=true,
    backref=true,
    firstinits=true,
]{biblatex}

\let \cite \parencite

\bibliography{sample}

\begin{document}
\title{Natural Language Processing for \\ 
Cognitive Analysis of Emotions}
%
%
\author{\textbf{Gustave Cortal}\inst{1} \and
\textbf{Alain Finkel}\inst{1, 4} \and
\textbf{Patrick Paroubek}\inst{2} \and \textbf{Lina Ye}\inst{3}}
\authorrunning{Cortal et al.}
%
\institute{Université Paris-Saclay, CNRS, ENS Paris-Saclay, Laboratoire Méthodes Formelles, 91190, Gif-sur-Yvette, France \and
Université Paris-Saclay, CNRS, Laboratoire Interdisciplinaire des Sciences du Numérique, 91400 Orsay, France \and 
Université Paris-Saclay, CNRS, ENS Paris-Saclay, CentraleSupélec, 91190, Gif-sur-Yvette, France \and
Institut Universitaire de France, France
\\
\email{gustave.cortal@ens-paris-saclay.fr}, \email{alain.finkel@lsv.fr}, \email{pap@limsi.fr}, \email{lina.ye@centralesupelec.fr}}
\maketitle              

\begin{abstract}

Emotion analysis in texts suffers from two major limitations: annotated gold-standard corpora are mostly small and homogeneous, and emotion identification is often simplified as a sentence-level classification problem. To address these issues, we introduce a new annotation scheme for exploring emotions and their causes, along with a new French dataset composed of autobiographical accounts of an emotional scene. The texts were collected by applying the Cognitive Analysis of Emotions developed by A.~Finkel to help people improve on their emotion management. The method requires the manual analysis of an emotional event by a coach trained in Cognitive Analysis. We present a rule-based approach to automatically annotate emotions and their semantic roles (e.g. emotion causes) to facilitate the identification of relevant aspects by the coach. We investigate future directions for emotion analysis using graph structures.

\keywords{Sentiment Analysis \and Aspect-Based Emotion Analysis \and Natural Language Processing \and Cognitive Analysis of Emotions \and Rule-Based system}
\end{abstract}

\section{Introduction}

\subsection{Cognitive Analysis of Emotions}

Similar to many psychological theories (e.g. Freud's psychoanalysis, Perls' Gestalt therapy, Greenberg's Emotion-focused therapy, Shapiro's Eye Movement Desensitization and Reprocessing and most psychological theories of emotions including Appraisal Theory), the Cognitive Analysis of Emotions (CAE)~\cite{Finkel2022} considers that the mind, in a given scene, processes emotions and associated cognitions according to a cycle. For the most part, this process is not conscious and begins with the identification of a situation and its issues. Then, it is followed by a reflection concerning the benefits and disadvantages of possible choices of actions. A decision is made, and the chosen action is executed. Finally, the cycle ends with a return to a ready state that is able to process the next scene. 

For instance, I am waiting for my turn to take a ticket at the cash desk of the cinema when someone passes me. I feel angry because I think neither I nor the social rules have been respected. I evaluate my possible actions and their consequences: protest verbally, physically push the person away, do nothing or run away. As my fear of a conflict overtakes my anger, I decide to keep quiet and do nothing. 

The conflict I avoided in the outside world may be internalized in my mind. I may be angry at the part of myself that didn't defend my rights, or I may be sad to be separated from my vision of a fair world. In this example, the emotion processing cycle did not go as well as possible. I remain mentally preoccupied after the scene. I have regrets and doubts. I mentally replay the scene differently. 

The CAE is part of discrete emotion theory as it studies how the four primary emotions (joy, sadness, anger and fear) appear in autobiographical accounts describing brief scenes (lasting a few tens of seconds) with emotions experienced by the author. One of CAE assumptions is that an emotion coming to our consciousness is a message to solve a problem (in the sense of problem-solving) associated with this emotion.

The universal problems, signaled by the four primary emotions, are formalized through the notion of territory. The previous scene stages an attack on the following two territories: my free time (constrained by the cinema queue) and my comforting vision of an organized and predictable world with laws accepted by almost everyone. Anger and fear are signals that at least one of our territories is under attack. It is up to us to defend (anger) or flee (fear) from it. Joy and sadness are signals that a change in our connection to a territory has occurred. For example, 
I may be joyful if I get a distinction in my master's degree because my important object \textit{positive self-image} will be reinforced. If I decide to work abroad, I may be sad because I will be physically separated from my important object \textit{family}.

Territories and objects are related to human needs that have to be satisfied. For example, according to Maslow's hierarchy of needs \cite{maslow:motivation}, human beings have physiological, safety, love and belonging, esteem, and self-actualization needs. In this paper, we propose to understand emotions and their causes by automatically identifying the relevant territories and objects involved in an emotional scene.

\subsection{Autobiographical accounts of an emotional scene}

In a CAE session, people who want to better manage their emotions write down an autobiographical narrative of a past emotional scene they experienced, in a given place and time, with identified characters. The coach imposes instructions for writing the scenes. These instructions represent textual metadata that will make easier the construction of an emotion analysis dataset. The author writes the account in four major parts:

\begin{itemize} \itemsep0em 
    \item \emph{Facts} describe the behaviors that are observable by everyone in the scene. This part also includes thoughts and physical feelings experienced by the author, because internal events are not observable but presumed \enquote{true}, as not refutable.
    \item \emph{Emotions} identify the emotions experienced by the author. Observable emotions of other participants can be considered as \textit{Facts}.
    \item \emph{Reasons} identify the emotion causes according to the territory theory of the CAE. Relevant territories and objects are identified.
    \item \emph{Actions} analyze the past actions, mentally replay the scene in the present and test possible actions for the future. The goal is to find the best actions adapted to the situation.
    
\end{itemize}

The CAE coach helps the author to identify, from the guided analysis of her or his account, the relevant territories and objects that are in play in an emotional scene. The coach's analysis aims to understand emotion causes and suggests corrective actions to better handle situations. In the next section, we describe the model we developed to automatically identify semantic roles (e.g. \textsc{experiencer}, \textsc{territory}, \textsc{object}, etc.) in a text. The proposed solution aims to automate an important step of CAE analysis, namely the identification of emotions and their causes.

\section{Emotion modelling based on Cognitive Analysis of Emotions}

\subsection{Sentiment and Emotion analysis}

Since its introduction by \textcite{PangLV02} two decades ago, sentiment analysis (a.k.a. opinion mining) has become an influential field of research with widespread applications in industry. However, the majority of research on sentiment analysis considers it as merely  text or content categorization task~\cite{sentiment}, i.e. classifying into two or three categories of sentiments: positive, negative, or neutral. In other words, sentiment analysis rarely takes into account the psychological aspect to really understand the sentiments and their causes. On the contrary, emotion detection aims at identifying distinct human emotion types expressed in texts, audio or videos \cite{8122047}. Besides studying the so-called primary 4-scale emotions, emotion detection also handles higher scale and even circumflex models, depending on both psychology theories and emotion models~\cite{SailunazDRA18}. A review of the existing annotated text datasets for emotion analysis has been done by \textcite{bostan-klinger-2018-analysis}. 

\subsection{French dataset for Aspect-Based Emotion Analysis\label{dataset}}

Some corpora with emotion annotation exist for French, e.g. the DEFT 2018 emotion, sentiment and opinion identification shared task dataset \cite{paroubek-etal-2018-deft2018} or the corpus for recognizing emotions in children's books \cite{etienne-etal-2020-lexpression}. However, they cannot be used for CAE for various reasons: the text material or the emotion model is incompatible, there is not enough data for model training, etc.

Our dataset, composed of autobiographical accounts of an emotional scene, will indicate emotions and their semantic roles: \textsc{cue} (a marker indicating the presence of an emotion, which can be a single word), \textsc{experiencer} (the author who feels an emotion), \textsc{target} (an entity or a person targeted by an emotion) and \textsc{cause} (an event that triggers an emotion). These roles are employed by \textcite{campagnano-etal-2022-srl4e} to unify several gold but heterogeneous datasets that contain annotations for both emotions and their semantic roles. Hence, instead of considering emotion analysis as a sentence-level classification problem, we focus on the aspect-level. We propose to deeply understand a given text describing an emotional scene, by automatically identifying who feels an emotion, what drives an entity to express an emotion toward a certain aspect and why. For instance, in this sentence, \enquote{Gustave loves carnivorous plants because they are beautiful}, Gustave (\textsc{experiencer}) exposes his joy (\textsc{cue}) towards carnivorous plants (\textsc{target}) because they are beautiful (\textsc{cause}).

\subsection{Extended scheme for emotion annotation}

We propose to extend the annotation scheme with new semantic roles based on CAE to better understand emotion causes. We introduce \textsc{territory} and \textsc{object}, corresponding to the notion of territory and object in CAE. We also introduce \textsc{attack} (expressions related to the act of attacking or being attacked, e.g. attack, assault, aggression, etc.) and \textsc{attacker} (an entity that attacks a \textsc{territory}). Identifying \textsc{attack} and \textsc{attacker} beforehand facilitates the identification of \textsc{territory}. For instance, in the sentence \enquote{My skills are attacked by Marc}, \enquote{My skills} are a \textsc{territory} related to the author's professional values and competent self-image that is attacked by the \textsc{attacker} \enquote{Marc}. These new semantic roles can be seen as a refinement of \textsc{cause} presented above. We also use two complementary roles: \textsc{modifier} for taking into account the intensity of an emotion (e.g. \enquote{I'm \textbf{a little} sad}) and \textsc{negation} to preserve the original meaning of expressions using negation markers (e.g. \enquote{She was \textbf{not} angry}). 

\section{Automatic identification of emotions and their semantic roles}
\subsection{Rule-based method}

We present a rule-based method to automatically identify the semantic roles in autobiographical accounts of an emotional scene. We leverage linguistic features using dependency parsing, co-reference resolution\footnote{\url{https://github.com/pandora-intelligence/crosslingual-coreference}} and part-of-speech tagging with the open source library SpaCy.\footnote{\url{https://spacy.io}}  Co-reference links are used to connect different expressions referring to the same referent, as it is useful to identify multiple occurrences of the same \textsc{experiencer} and the same \textsc{target} to better understand the emotional flow in a text.   

We use WordNet \cite{wordnet}, a lexical database of semantic relations including synonyms, hyponyms, and meronyms, to identify \textsc{cue}s and words related to an \textsc{attack} of a \textsc{territory}. For the French language, we choose the French WordNet called WOLF \cite{wolf}. Sentiment and emotion lexicons are also used to improve the identification of \textsc{cue}s. SentiWordNet \cite{baccianella-etal-2010-sentiwordnet} is built on top of WordNet. In this lexicon, each word sense is assigned with a degree of positivity, negativity, and neutrality. NRC Emotion Lexicon \cite{nrcemotion} is another popular lexicon, where each word is associated with eight basic emotions (anger, fear, anticipation, trust, surprise, sadness, joy, and disgust) and two polarities (negative and positive).   

As we are working with autobiographical texts, an author often describes oneself with first-person pronouns. It is therefore easy to detect the \textsc{experiencer} through regular expression filtering. To identify complex semantic roles such as \textsc{territory}, we manually define several rules using linguistic features. For example, a \textsc{territory} is found if it is the subject of an \textsc{attack} in passive voice: \enquote{\textbf{My skills} are attacked by Marc}.

Rule-based methods do not require training data. Explainability of results is one of its major benefits. However, it is sometimes difficult to formulate rules and a task may require a huge amount of rules, leading the method to be highly domain-specific. Coherence and consistency checking time increase drastically with the number of rules. Performance stability on yet unseen data is difficult to assess. At the time of writing, the annotation of the autobiographical accounts has not yet been performed, nor has the evaluation of this first rule-based prototype. This will be done in the near future.

In future work, we will combine our rule-based method with recent deep learning techniques to take the best of both approaches. For example, \textcite{mixed} propose to combine a rule-based method with a deep convolutional neural network to improve the performance of aspect extraction. \textcite{li-etal-2021-weakly} show that neural taggers can generate new rules based on seed rules, which are manually predefined high quality rules. For our case, the learned rules can potentially explain the predicted semantic roles, and hence discover new ways to identify the emotional content of a text.  

\subsection{Graph structure}

\begin{figure}[!htp]
    \centering
    \includegraphics[scale=0.17]{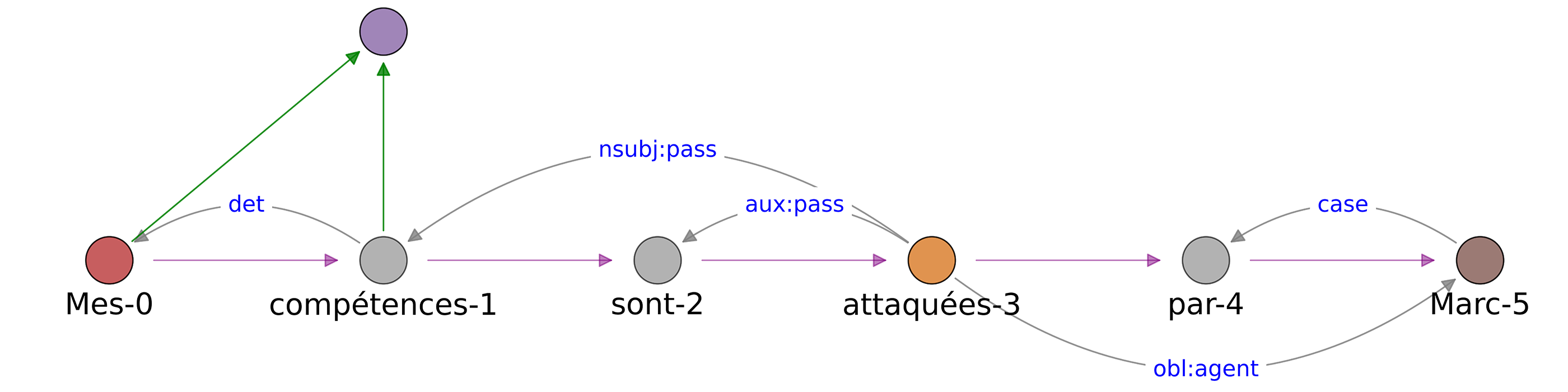}
    \caption{Visualization of the French sentence: \enquote{\textit{Mes compétences sont attaquées par Marc}} (translated \enquote{My skills are attacked by Marc}). \textbf{Edge colors} indicate different types of relations, e.g. noun chunk membership is in green and sequential relation is in pink (e.g. from \textit{Mes-0} to \textit{compétences-1}). \textbf{Node colors} indicate semantic roles, e.g. \textsc{experiencer} is in red (\textit{Mes-0}), \textsc{territory} is in purple (\textit{Mes-0 compétences-1}), \textsc{attacker} is in brown (\textit{Marc-5}) and \textsc{attack} is in yellow (\textit{attaquées-3}).}
    \label{fig:viztool}
\end{figure}

An emotion is a complex phenomenon that resonates in multiple levels of analysis through different scales. We propose to represent emotion expressions by a graph structure that can be visualized. A sentence or a whole text corresponds to a graph in which nodes are words and edges indicate relations of various kinds between words. We incorporate our rule-based method into the graph structure. Figure \ref{fig:viztool} illustrates the visualization application built using NetworkX\footnote{\url{https://networkx.org}}. The application can display different levels of text analysis (e.g. dependency parsing, our emotion analysis, etc.) in a single plane. For instance, co-reference links connect certain semantic roles between them. The relation visualizer can therefore be used to facilitate the manual process of creating rules.

We plan to augment the graph structure with new semantic relations by extracting knowledge paths from ConceptNet \cite{conceptnet}. It is a multilingual semantic network that provides concepts connected with large amounts of semantic relations. For instance, \textcite{yan2021position} incorporate commonsense knowledge from ConceptNet to reduce the position bias in Emotion Cause Extraction models. We believe semantic networks can be useful to better capture the dependencies between emotions and their semantic roles, as they leverage commonsense knowledge. Our next goal is to design dedicated graph neural networks such that the specific structural elements of these graphs can be better captured to improve performance. For example, \textcite{marcheggiani-titov-2017-encoding} exploit syntactic information using graph convolutional networks to encode sentences, as semantic representations are similar to syntactic ones. The proposed methods can be combined and extended to exploit other information that the graph structure offers.

\section{Conclusion}

To remedy some limitations in emotion analysis, we propose to deeply understand an emotional scene by performing fine-grained analysis of an emotion and its semantic roles at the aspect-level. We introduce a new annotation scheme, based on Cognitive Analysis of Emotions, along with a new dataset composed of French autobiographical accounts of an emotional scene. As manually analyzing accounts is time-consuming, we provide an automated assistant for the coach, who can therefore focus on aspects that cannot be submitted to automatic processing. Our rule-based method automatically identifies emotions and their semantic roles in a text. In the future, after annotating the autobiographical accounts with our annotation scheme and performing the quantitative evaluation of our rule-based method, we plan to combine it with recent deep learning models (e.g. graph neural networks) through the graph structure we developed to improve performance. 

\newpage

\newpage
\printbibliography 

\end{document}